\documentclass[a4paper]{article}

\usepackage{amsthm,amsmath,amsfonts,amssymb}
\usepackage{color}

\newcommand{\F}{\mathcal{F}}

\newcommand{\PA}{\mathcal{P}}
\newcommand{\R}{\mathcal{R}}

\newcommand{\ep}{\epsilon}

\newcommand{\Z}{\widetilde{Z}}
\newcommand{\p}{\tilde{p}}
\newcommand{\SA}{\mathcal{S}}

\theoremstyle{plain}
\newtheorem{theorem}{Theorem}[section]

\theoremstyle{definition}

\theoremstyle{remark}

\numberwithin{equation}{section}

\providecommand{\keywords}[1]
{
  \small	
  \textbf{\textit{Keywords---}} #1
}

\begin{document}

\title{Stochastic approximation  in non-markovian environments revisited\thanks{Work  supported by  a National Science Chair from Government of India. The author thanks Chiranjib Bhattacharya and Bruce Hajek for useful discussions.}}

\author{Vivek S.\ Borkar \\ Department of Electrical Engineering,\\ 
Indian Institute of Technology Bombay,\\
Powai, Mumbai 400076, INDIA.\\ 
E-mail: borkar.vs@gmail.com 
}

\maketitle

\begin{abstract}
Based on some recent work of the author on stochastic approximation in non-markovian environments, the situation when the driving random process  is non-ergodic in addition to being non-markovian is considered. Using this, we propose an analytic framework for understanding transformer based learning, specifically, the `attention' mechanism,  and continual learning, both of which depend on the entire past in principle.
\end{abstract}


\bigskip

\section{Introduction}

In \cite{B-NM}, it was shown that stochastic approximation  with stationary non-markovian noise behaves like stochastic approximation with a stationary Markov noise with identical pair marginals, modulo an additional asymptotically vanishing error term. In this sequel, we consider the situation where the original non-markovian noise is not ergodic and argue that its ergodic decomposition gets reflected in the Doeblin decomposition of the state space for the equivalent Markov noise. In particular, this implies that the iterates retain some memory of the distant past, as reflected in the tail $\sigma$-field at $-\infty$. We argue that this in principle leads to a paradigm for long-term memory in machine learning and interpret transformers and continual learning in this light.\\

The next section recalls in brief the background for stochastic approximation with Markov noise, followed by the results of \cite{B-NM} in section 3, leading to our main result. Section 4 sketches some potential implications for the attention mechanism in machine learning. It also discusses connections to continual learning and a few other themes. 
Given the enormous current interest in this topic, the latter themes are being studied from multiple angles, see, e.g., \cite{Chang}, \cite{Goel} to quote two recent contributions. The approach here is, however, quite distinct.\\

\section{Stochastic approximation with Markov noise}

A generic stochastic approximation scheme (SA for short) in $\R^d$ is given by
\begin{equation}
x(n+1) = x(n) + a(n)\Big(h(x(n), Z(n)) + M(n+1)\Big), \ n \geq 0, \label{SA}
\end{equation}
where 
\begin{enumerate}

\item $h : \R^d \to \R$ is Lipschitz,\\

\item $M(n), n \geq 0,$ is a \textit{martingale difference sequence}, i.e., for $\F_n :=$ the $\sigma$-field $\sigma(M(k), Z(k), a(k), k \leq n; \ x(0))$, $n \geq 0,$ we have
$$E\left[M(n+1) | \F_n\right] = \theta \ \ (:= \ \mbox{the zero vector)},$$
satisfying the moment bound
$$E\left[\|M(n+1)\|^2 | \F_n\right] \leq K(1 + \|x(n)\|^2) , \ n \geq 0,$$

\medskip

\item $Z(n), n \geq 0$, is the so called \textit{Markov noise}, i.e., a process taking values in a Polish space\footnote{i.e., separable with a compatible metric that is complete. This is a convenient level of generality because most spaces in applications, such as the Banach space of continuous functions on $[0,T]$ that arises in diffusion models, are Polish, and because most of the classical results for probability measures on $\R^d$ carry over to probability measures on Polish spaces.}, such that there exists a (parametrized)  transition probability kernel
$$p_x(ds'|s) : (s,x) \in S\times\R^d \mapsto \PA(S)$$
which satisfies:

\begin{enumerate}
\item it is Lipschitz in $x$,

\medskip

\item $P(Z(n+1) \in A|\F_n) = p_{x(n)}(A|Z(n)), \ n \geq 0,$

\medskip

\item for fixed $x$, it has a unique stationary distribution $\pi_x(ds) \in \PA(S)$.
\end{enumerate}

\medskip

\item the step size sequence $a(n), n \geq 0$, possibly random, satisfies the \textit{Robbins-Monro} conditions
$$\sum_na(n) = \infty, \ \sum_na(n)^2 < \infty.$$

\end{enumerate}

\medskip

For simplicity, we shall take $S$ to be compact. This avoids making assumptions about and keeping track of various moments of continuous $\R^d$-valued functions of $S$-valued random variables.\\

We shall follow the `ODE' (for \textit{Ordinary Differential Equations}) approach to analyze SA. Consider the ODE
\begin{equation}
\dot{y}(t) = \int h(y(t), z)\pi_{y(t)}(dz). \label{ODE}
\end{equation}
Under our assumptions, the map $x \in \R^d \mapsto \int h(x, z)\pi_{x}(dz)$ is Lipschitz. For simple $S$ (e.g., finite dimensional or discrete), simple sufficient conditions can be given for this to hold.\\

 In particular, the Lipschitz property ensures that \eqref{ODE} is \textit{well-posed}, i.e., it has a unique solution in both forward and backward time for every choice of $x(0)$, and this solution depends continuously on $x(0)$ as a map from $\R^d$ to $C((-\infty, \infty); \R^d) :=$ the space of continuous functions $(-\infty,\infty) \to \R^d$ with topology of uniform convergence on compacts. \\
 
 A set $B \subset \R^d$ is said to be invariant for \eqref{ODE} if the unique trajectory $y(t)$ through any $x \in B$ remains in $B$ for $-\infty < t < \infty$. It is said to be \textit{internally chain transitive} if for any given $\ep, T > 0$, $n \geq 2$, and $x_1, \cdots, x_n \in B$, there exist trajectories of duration at least $T$ that begin in the $\ep$-neighbourhood of $x_i$ and end in the $\ep$-neighbourhood of $x_{i+1}$. One then has the following result  due to Benaim \cite{Benaim}.

\begin{theorem}\label{Benaim} Almost surely, $x(n) \to$ an internally chain transitive invariant set of \eqref{ODE}. \end{theorem}

\section{Non-markovian environment}

Now consider the case when $\{Z(n)\}$ is not a Markov noise, but an arbitrary random process. This is precisely the situation studied in \cite{B-NM} and some of this section will be devoted to recalling the results of \cite{B-NM} in the present context. Fix $x \in \R^d$. Define the \textit{Markov mimic}\footnote{See \cite{ABG} for an extensive account of this concept.}  $\{\Z(n)\}$ of $\{Z_n\}$  as the $S$-valued, possibly time-inhomogeneous,  Markov noise defined by the parametrized transition kernel
\begin{eqnarray*}
\p_{x,n}(A|z) &:=& P(\Z(n+1) \in A | \Z(n) = z, x(n) = x) \\
&:=& P(Z(n+1) \in A | Z(n) = z, x(n) = x), \ n \geq 0.
\end{eqnarray*}
Fix $x \in \R^d$. Let $S_i, -\infty < i < \infty$, denote replicas of $S$ and define
$$\SA^- := \prod_{i=-\infty}^0 S_i, \ \SA^* := \prod_{i=-\infty}^\infty S_i,$$
Then, setting $Z(n) =$ a fixed element $s^* \in S$ for $n < -N$,  $Z^-(n) := [Z(n), Z(n-1), \cdots], -\infty < n < \infty$, is trivially an $\SA^-$-valued Markov chain. Its transition probability is completely specified by the one step conditional law
$P(Z(n+1)|Z(k), k \leq n)$, which we take to be given by a fixed map
$$z^\infty \in \SA^- \mapsto p_x(dz|z^\infty) \in \PA(S).$$
We assume that this map is continuous in $(x,z^\infty)$.\\

 Since $\SA^-$ is compact by Tychonoff's theorem, standard theory of Markov chains on general spaces  (see, e.g., \cite{MT}) ensures the existence of a convex compact nonempty set of invariant probability measures, the extreme points of which correspond to ergodic processes. In particular, each invariant measure leads to a stationary process $\{Z(n)\}$ on the entire time axis $-\infty < n < \infty$. The corresponding $\{\Z(n)\}$ will then be a time-homogeneous Markov chain on $S$ with a time-independent transition kernel $p_x(dz'|z), z \in S$, parametrized by $x \in \R^d$. Furthermore, the ergodic decomposition of invariant measures for $\{Z(n)\}$ leads to an identical ergodic decomposition for $\{\Z(n)\}$. Since $\{\Z(n)\}$ is a time-homogeneous Markov chain, this corresponds to a decomposition of its state space into a.s.\ disjoint ergodic classes (the Doeblin decomposiion) and leads to a compact convex nonempty set of invariant measures for the Markov chain $\{\Z(n)\}$ whose extreme points are supported on the component sets of the Doeblin decomposition. These correspond to ergodic Markov chains.\\

The main result of \cite{B-NM} is that when one performs the iteration \eqref{SA} with stationary non-markovian $\{Z(n)\}$, one is equivalently running \eqref{SA} for its Markov mimic $\{\Z(n)\}$, modulated by $\{x(n)\}$. There is simply an additional asymptotically negligible error term due to non-markovianity. The reader is referred to \cite{B-NM} for details.\\

The main takeaway from \cite{B-NM} in the present context then is that \eqref{SA} is tracking the ODE
\begin{equation}
\dot{z}(t) =  \int h(z(t), z)\psi_{z(t)}(dz), \label{ODE2}
\end{equation}
where $\psi_x$ is one of the extremal invariant measures corresponding to an ergodic class. 
Recall that the time index is $-\infty < n < \infty$. The information regarding which precise $\psi_x$ is operative will be contained in the `tail $\sigma$-field at $-\infty$', i.e.,
$$\sigma_{-\infty} \ := \ \cap_{n = -\infty}^\infty\sigma\left(\Z(k), k \leq n\right).$$
This $\sigma$-field need not be countably generated, i.e., need not be representable as the smallest $\sigma$-field containing a prescribed countable family of sets. However, for illustrative purposes, let us assume it to be \textit{finitely} generated, i.e., generated by a finite partition $A_1, \cdots , A_M$ of $S$ that is independent of $x$. Then each $A_i$ corresponds to a single ergodic class, implying a unique $\Psi_x$. The process thus carries the memory of which ergodic class it started in. The finite ergodic decomposition was purely for expository convenience. There can be infinitely many, possibly uncountable, ergodic classes, suggesting that there is no a priory limitation on how much of the past can be encoded in an ergodic decomposition.
\\

One can be more specific. Suppose the transition probabilities of $\{Z_n\}$ are of the form
\begin{equation}
P(Z_{n+1} \in A | \F_n) = \check{p}_{x(n)}(A|Z_n, g(Z_0, \cdots, Z_k)), \ n \geq k,  \label{template1}
\end{equation}
for some fixed $k \geq 0$, a measurable $g: S^{k+1} \to \R$, and a suitable transition probability function $\check{p}(\cdots|\cdot, \cdot)$. This amounts to the conditional law of the past at time $n$ given $X_n$ being a Dirac measure at $g(Z_0, \cdots, k)$ for all $n > k$, Then in the $n \to \infty$ limit, the ergodic classes will be identified with the distinct values that the random variable $g(Z_0, \cdots, Z_k)$ takes.\\

Extending this idea, consider an increasing sequence of a.s. finite $\{\F_n\}$-stopping times $\tau_n, n \geq 0,$ with $n \uparrow \infty$, and  $\R^\ell$-valued random variables $\zeta_n, n \geq 0,$ adapted to the stopped $\sigma$-fields $\F_{\tau_n}, n \geq 0$. Let $\alpha \in (0,1)$. Assuming for simplicity that $\{\|\zeta_n\|\}$ have a common deterministic bound a.s., the random variables 
$$\xi(n) := \sum_{m=0}^n\alpha^{\tau_m}\zeta_m, \ n \geq 0,$$ 
are well defined.
Consider then  the non-markovian kernel
\begin{equation}
P(Z_{n+1} \in A | \F_n) = \check{p}_{x(n)}(A|Z_n, \xi(n)), \ n \geq 0. \label{template1}
\end{equation}
This can model continual learning. The process will be asymptotically stationary if $\{\xi_n\}$ are so and the structure of the latter will decide the ergodic classes. Note the subtle point that the conditional law of the past given $X_n$ at time $n$ has to satisfy some natural consistency conditions as $n$ varies. These come easy for the exponentially decaying weights as above.\\

This plays a role when applied to machine learning in non-markovian environments. We discuss this next. 

\section{Non-markovian SA in machine learning}

Coming back to our original motivation of application of SA in non-markovian environments that may arise in machine learning applications, what this means is that a sample trajectory of the SA started at $-\infty$ will have encoded in its evolution the information of its ergodic class. Once again, assume that the tail $\sigma$-field is generated by a finite partition of $S$ that is independent of the process $\{x(n)\}$. What this means is that some essential `\textit{feature(s)}' are being hard coded into the dynamics. Since in practice we do not start at $-\infty$, this will be an emergent phenomenon as $n\uparrow\infty$, its impact in what is observed being a function of the rate of convergence to ergodicity. This is dictated by the spectral gap for the Markov chain $\{\Z(n)\}$ \cite{MT}.\\

A prominent example of non-markovianity in machine learning occurs in the structure of an ideal  transformer, where the `\textit{attention}' mechanism depends on the entire past. This renders everything built upon transformers, such as LLMs, a non-markovian object at some stage of its training. In reality, the transformer does not involve the entire past, but a large window into the past. Nevertheless, viewing large time windows into the past as an approximation to the infinite past, the above picture gives some intuition about the transformer's attention mechanism zeroing on the relevant aspects of the past. \\


Another emerging application is that of elephant random walks, a model from statistical physics, that lends itself to some analysis as non-markovian stochastic approximation \cite{Das}, \cite{Podder}.\\

An added complication is the slow modulation of $\{Z_n\}$ by the iterates $\{x(n)\}$ moving on a slower time scale. This implies that the ergodic decomposition will itself change slowly, possibly in a discontinuous manner. In the asymptotic regime as $n\to\infty$, if $x(n)$ converges, then we are essentially back to the above framework, with a \textit{random} ergodic decomposition in case $\lim_{n\to\infty}x(n)$ is random.\\

Consider the simplest and the most popular special instance of the foregoing, i.e.\ the stochastic gradient descent. Here $h(x,u) = -\nabla_xF(x,u)$ where $\nabla_x$ is the gradient in the $x$ variable, $F: \R^d \to \R$ in the simplest case may be taken to be  twice continuously differentiable, and $\lim_{\|x\|\to\infty}F(x) = \infty$ with at most quadratic growth. Suppose in addition that it has only finitely many local minima. The ODE now is
$$\dot{x}(t) = -\int\nabla_xF(x(t),z)\pi_{x(t)}(dz).$$
One critical aspect of this equation that is often not noticed in literature is that in general,
$$\int\nabla_xF(x,z)\pi_x(dz) \neq \nabla\left(\int F(x,z)\pi_x(dz)\right).$$
However, in practice one uses not the exact gradient, but a finite difference based approximation of it, such as the Kiefer-Wolfowitz scheme \cite{KW}, Simultaneous Perturbation Stochastic Approximation (SPSA) \cite{Spall}, the Mukherjee-Wu-Zhao scheme \cite{Mukherjee}, the Flaxman-Kalai-McMahan scheme \cite{Flaxman}, etc. These approximate the r.h.s.\ rather than the l.h.s., so it works out fine.\\

Non-markovianity in control applications is real, because a Markov model, controlled or otherwise, is usually an approximation at best. Thus the above analysis of SA in a non-markovian set up is very much relevant for schemes such as policy gradient algorithms. See, e.g., \cite{Ch} for non-markovianity in reinforcement learning schemes.\\


Another important application domain is when one approximates the state space of the process by discretization or by using features, losing thereby its markovianity. This is considered in \cite{B-NM}.
What is new in the present work is the next step to the ergodic decomposition and its implications.

\end{document}